\documentclass{article}
\usepackage{ijcai22}
\usepackage[para]{footmisc}
\usepackage{times}
\usepackage{soul}
\usepackage{url}
\usepackage[hidelinks]{hyperref}
\usepackage[utf8]{inputenc}
\usepackage[small]{caption}
\usepackage{graphicx}
\usepackage{amsmath}
\usepackage{amsthm}
\usepackage{booktabs}
\usepackage{algorithm}
\usepackage{algorithmic}
\usepackage{color}
\title{A Survey on Semantics in Automated Data Science}

\author{
    Udayan Khurana \and 
    Kavitha Srinivas \and 
    Horst Samulowitz
    \affiliations
    IBM Research AI 
    \emails
    \{ukhurana,kavitha.srinivas,samulowitz\}@us.ibm.com 
}

\begin{document}

\maketitle

\begin{abstract}
  Data Scientists leverage common sense reasoning and domain knowledge to understand and enrich data for building predictive models. In recent years, we have witnessed a surge in tools and techniques for {\em automated machine learning}. While data scientists can employ various such tools to help with model building, many other aspects such as {\em feature engineering} that require semantic understanding of concepts, remain manual to a large extent. 
  In this paper we discuss important shortcomings of current automated data science solutions and machine learning. We discuss how leveraging basic semantic reasoning on data in combination with novel tools for data science automation can help with consistent and explainable data augmentation and transformation. Moreover, semantics can assist data scientists in a new manner by helping with challenges related to {\em trust}, {\em bias}, and {\em explainability}.
\end{abstract}

\section{Introduction}
In recent years, the automation of machine learning (AutoML) or data science processes has received considerable attention\footnote{\url{ https://www.tinyurl.com/forbes-automated-data-science/}\hspace{0.2cm}}.
The primary driving force behind it is the desire to reduce human intervention and thereby save time and cost, and improve efficacy of the modeling process. Such automation has been achieved through various techniques such as reinforcement learning based agents, multi-arm bandits, numerical optimization methods such as bayesian optimization and others. 
While the automation efforts proposed so far have led to efficiencies in different parts of the data science (DS) process, one dimension that seems to be relatively untouched is that of semantics in data. Traditionally in data science, where the process is heavily dependent on a human data scientist or a domain expert\footnote{We refer to a data scientist as the conductor of ML or DS process.}, semantics of the data plays an important role. The data scientist studies the given problem or data and relates them to the concepts in the real world and uncovers relationships between them. This is followed by linking the problem, typically a regression or a classification task defined by the target variable, to the concepts represented by features. Finally, the data scientist utilizes the knowledge of the domain, or general knowledge of the world to perform operations which enhance the modeling capability of the given data, such as feature engineering (FE). 
Amongst the automation techniques thus far, only few have made progress concerning semantics in data science. Much of this area is still open for further progress which is much needed in industry and is of interest to research as well. In this paper, we survey the work done so far in semantic data science. We also survey the rest of automated data science from the perspective of extending those techniques and methods in a semantic direction.

\begin{figure}
    \centering
    \includegraphics[scale=.25]{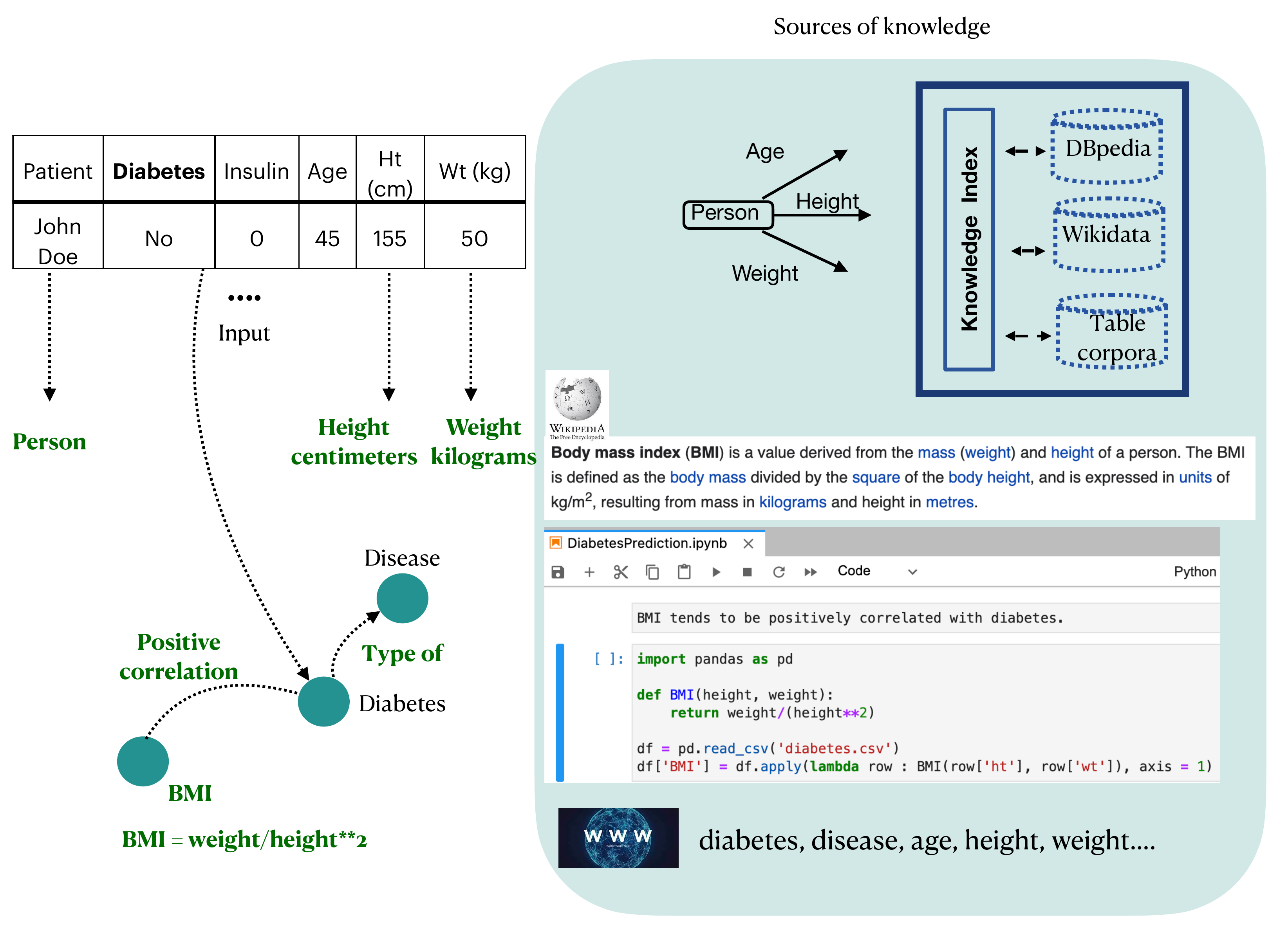}
    \caption{Motivating example on modeling Diabetes with Semantics}
    \label{fig:example}
\end{figure}

Figure~\ref{fig:example} shows a motivating example for injecting semantics into data science.  The input is a table about modeling diabetes as a function of age, height, weight, etc.  It is now possible (albeit not perfectly) to imbue this data with semantics shown in green labels to help make sense of the data. The right blue panel shows the sources of knowledge available that can be leveraged to add semantics.  Knowledge graphs such as DBpedia, Wikipedia, or massive corpora of tables on the web can help make sense of columns.  As an example, column to concept mapping technology~\cite{khurana2020semantic} can help map named entity columns such as ``Patient'' to the concept \texttt{Person}.  The Person concept in knowledge graphs frequently has a set of attributes in knowledge graphs, which can be exploited to match ``Ht'' to \texttt{Height}, using approaches for column to property matching.  Units can be approximated by signatures on various attribute distributions in knowledge graphs.  In addition to knowledge graphs, there is a wealth of information buried in unstructured sources such as Wikipedia, text cells of data science notebooks and code. Mining techniques on these can be used on these corpora to automatically create features for the data science problem at hand.  As an example, the formula for calculating BMI can be mined from both code and text in Wikipedia, and BMI can be used as an semantics-based feature to improve diabetes prediction.  Furthermore, formulae mined from notebooks can either allow for the extraction of expressions for feature engineering, or provide hints on when such augmentation is likely to produce a meaningful signal, for instance, if the expression appears often across notebooks working with similar data. 

Feature engineering is central to any successful model building task. It is the process of altering the feature space representation to better suit model fitting or predictability of the target variable~\cite{fechapter}. It is also a step that heavily involves semantics. One example is illustrated in the BMI derivation in Figure~\ref{fig:example}, where we outline how an automation of this process can work via Wikipedia.  To show the generalizability of using semantics for feature engineering, we add two more examples drawn from different domains.  First, consider an example of loan approval prediction from Kaggle\footnote{\url{https://www.tinyurl.com/yonatanrabinovich-loan}}
where the goal is to predict whether a given loan application will be approved or not. Amongst the provided attributes are the loan amount, the gender of the applicant, marital status, education, applicant income, co-applicant income and others. A data scientist knowledgeable about the domain of loans would arrive at ${applicant\_income + coapplicant\_income} \over {loan\_amount}$ as one of the predictors. Performing this step of feature engineering requires various levels of semantic operations. This includes primarily mapping given columns to real-world concepts; then, finding relationships between these feature concepts; finally, finding which relationships make sense with respect to the target concept and so on.

Our second example is drawn from Kaggle datasets on COVID\footnote{\url{https://www.kaggle.com/imdevskp/corona-virus-report}}.  A common ratio that is used for modeling the disease is the case fatality rate (CFR).  CFR is a formula relating the number of deaths to confirmed cases such that ${\text{CFR}}{\%}={\frac {\text{Number of deaths from disease}}{\text{Number of confirmed cases of disease}}}\times 100$.   Such formulas can be found in both code manipulating COVID notebooks as well as in Wikipedia, and again can be used to create new features for model building.

In addition to feature engineering, other aspects of machine learning (ML) and data science are also dependent on semantic interpretation of data and models. For example, model explainability is based in part on interpretation of concepts in the data and the meaning of operations performed on them. In order to ensure fairness through de-biasing, we require recognition of traditional points of bias~\cite{mehrabi2021survey} and ensuring the same are not repeated in the future. In order to automate this process, semantic concept recognition would be central to any meaningful system that identifies and removes it. Similarly, enforcing business rules through symbolic-AI~\footnote{\url{https://tinyurl.com/inrule-business-rules-ml}}
and semantics enhances the case for semantically driven automated data science.  In addition, initial work has targeted understanding hyper-parameters and extraction of constraints in applying them \cite{hpodoc}, as well as use of semantics for data cleansing \cite{6982731}.  

In this paper, we first survey automated machine learning at a high level (Section~\ref{automl_section}). We then focus our attention on automated feature engineering (Section~\ref{autofe_section}) with the popular non-semantic work and the recent developments in semantics-oriented feature engineering. We conclude that section by suggesting our vision for introducing a greater level of semantics to the popular feature engineering techniques.
Then, we talk about relevant works in semantic understanding and annotation of data (Section~\ref{data_section}), code (Section~\ref{code_section}) and text (Section~\ref{nlp_expression_section}) understanding respectively, as the means of building infrastructure for semantics-oriented data science. We conclude with extending these ideas for trust, bias and explainability in machine learning from a semantic dimension (Section~\ref{misc_section}). 

\section{Overview of Automated Data Science}
\label{automl_section}
In existing work on AutoML, Auto-sklearn~\cite{feurer-neurips15a} and Auto-WEKA~\cite{autoweka2} use sequential parameter optimization based on Bayesian Optimization to determine effective predictive modeling pipelines by combining data pre-processors, transformers and estimators. Both variants are based on the general purpose algorithm configuration framework SMAC~\cite{smac} to find optimal machine learning pipelines. In order to apply SMAC, the problem of determining the appropriate ML approach is cast into a configuration problem where the selection of the algorithm itself is modeled as a configuration.
Auto-sklearn also supports warm-starting configuration search by trying to generalize configuration settings across data sets based on historic performance information.  It also constructs an ensemble of classifiers instead of a single classifier. It uses ensemble selection from~\cite{ensembleselection} which is a greedy algorithm that starting from an empty set of models incrementally adds a model to the working set in each step if it results in maximizing the improvement of predictive performance of the ensemble. 

Another approach for automated ML is based on genetic algorithms~\cite{olson2016tpot}. While it does not create ensembles, it could compose them based on derived models as the authors point out. More interestingly, the approach presented in~\cite{geneticensembles} uses multi-objective genetic programming to evolve a set of accurate and diverse models via biasing the fitness function accordingly.
APRL~\cite{aprl} is based on an agent through reinforcement learning that uses a reward function related to model accuracy through ensemble generalization error. It mimics the trial and error approach of a data scientist. 

While the above works focus on solving the entire AutoML problem as one, there are also a variety of methods and techniques have been proposed to automate different parts of the process. 
Some examples are hyper-parameter configuration
~\cite{hpo}, feature engineering~\cite{fechapter}, model selection~\cite{sabharwal2016selecting} and data cleaning~\cite{krishnan2019alphaclean}. 
A recent survey by Zoller et al.~\cite{zoller2021benchmark} lists various techniques for different components of the machine learning cycle.
Similarly, vast amount of work exists in automated ML for neural networks also leveraging approaches from optimization, reinforcement learning, and genetic algorithms~\cite{he2021automl}.

While we focus mostly on the impact of semantics on automated data engineering and understanding, it can also be leveraged as part of the AutoML optimization itself. For instance, within hyper-parameter optimization, we may analyze code and Python DocStrings to not only determine available hyper-parameter settings automatically~\cite{hpodoc}, but also to extract constraints on hyper-parameter configurations such as mutually exclusive settings. This at least reduces the search space to valid choices automatically. 
At the pipeline configuration level, one can observe patterns in existing code such as PCA being nearly always preceeded by normalization and not the other way around and use them for instance as a way to prioritize specific pipelines in the automated pipeline exploration.

\section{Automated Feature Engineering}
\label{autofe_section}
In this section we delve deeper on automations in feature engineering (AutoFE). We first capture the work that ignores semantics of the data and discuss how there is a missed opportunity in incorporating semantics. Next, we talk about some recent research that is based on semantics. Finally, we present our ideas with suggestions to extend various semantic-oblivious approaches with semantics.
\subsection{Semantics Oblivious Approaches}
\label{autofe_nonsem}
There is a diverse set of approaches towards automated feature engineering which are summarized below.
FICUS~\cite{MarkovitchRosenstein02} performs a beam search over the space of possible features, constructing new features by applying constructor functions. It is guided by heuristic measures based on information gain in a decision tree.
Data Science Machine by~\cite{kanter:dsm} applies all transformations on all features at once (but no sequence of transformations), then performing feature selection and model hyper-parameter optimization over the augmented dataset. FEADIS~\cite{DorReich12} works through a combination of random feature generation and feature selection. 
ExploreKit~\cite{DBLP:conf/icdm/KatzSS16} expands the feature space explicitly, one feature at a time. It employs learning to rank the newly constructed features and evaluating the most promising ones. 
LFE~\cite{lfe} directly predicts the most useful transformation per feature based on learning effectiveness of transforms on sketched representations of historical data through a perceptron. 
The approach introduced in~\cite{recommendersystem2018} is fully focused on leveraging historical information using a recommender system. The approach is very effective in determining a pipeline while being limited by only being able to select from a fixed set of pre-existing pipelines.
More recent and initial work~\cite{alphad3m} inspired by AlphaGo Zero~\cite{alphazero} models pipeline generation as a single-player game and builds pipelines iteratively by selecting among a set of actions which are insertion, deletion, replacement of pipeline parts.
Cognito~\cite{khurana2018feature,khurana2016cognito} use reinforcement learning to optimize feature transformation through an agent.

\subsection{Missed Opportunities due to lack of Semantics in Automated Feature Engineering}
The state-of-the-art feature engineering systems mentioned above rely on trial and error for generating new features through transformation functions on the provided features. They try to maximize accuracy but are oblivious of the actual semantics of the data and transformation functions. This has several drawbacks. 
Firstly, they can not utilize external knowledge to perform specific transformations on data which might be very difficult to stumble upon using generalized exploration techniques. For example, BMI (Figure~\ref{fig:example}) is a complex formula which any of these techniques will take a very long time and number of trials before discovering. Secondly, they could violate physical laws such as addition of {\texttt height} and {\texttt weight}, which would be a valid exploration option in these methods but through worldly knowledge, we know that it does not physically make sense and should be avoided. Third, these methods do not have the capability of bringing additional data not provided in the problem but available externally and help improve the modeling capacity. 
Existence of abundance of knowledge in the form of knowledge bases, digital books, wikipedia, blogs and so on is a missed opportunity for automated feature engineering.

\subsection{Semantic Approaches}
There is recent work on enhancing a given dataset with new columns that help make more accurate predictive models. Friedman et al.~\cite{friedman2018recursive} and Galhotra et al.~\cite{galhotra2019automated} recursively generate new features using a knowledge base for text classification and web tables, respectively. FeGeLOD~\cite{PH} extracts features from Linked Open Data via queries. These approaches, however, fail to thoroughly explore the available information from multiple data sources. S3D~\cite{s3d} is able to bring in data from multiple sources such as webtables and multiple knowledge graphs but adds new columns irrespective of the target variable. Semantic Feature Discovery~\cite{sfd} uses formulae from a historical repository of data science code files and notebooks by linking concepts in a given data problem with those in the historical code. By reusing these formulae, it generates potentially useful new features. While these approaches are promising, their impact is bounded due to the limited availability and richness of knowledge bases, and the quality of underlying concept matching algorithms. However, we expect that to improve over time.

\subsection{Proposal for Utilizing Semantics in AutoFE}
~\cite{cherrier2019consistent} use a genetic programming formulation for feature construction, but with constraints in the domain of physics. Examples of constraints are -- energy and dimension can not be added as it does not make sense physically, and performing square after square root does not make much sense either, and so on. We argue a generalization of this approach on top of existing non-semantic approaches can significantly benefit latter's efficacy. Whether it is feature engineering through genetic programming as advised by~\cite{cherrier2019consistent} or through other optimization methods described in Section~\ref{autofe_nonsem} such as Cognito~\cite{khurana2018feature}, DSM~\cite{kanter:dsm}, ExploreKit~\cite{DBLP:conf/icdm/KatzSS16} or even metalearning approach such as LFE~\cite{lfe}, by identifying semantic types and enforcing certain constraints (some of which one can argue are global, whereas some specific to the problem) that can be expressed in an appropriate grammar are beneficial. Any of these methods can be restricted in their (non-beneficial) operations by simply checking for semantics or constraint consistency. 
This would require a common grammar to represent such constraints and an effective engine to enforce them. It might of interest for researchers in this area to develop a language that is expressive enough to capture constraints from various domains but simple enough to specify them easily or translate existing business or scientific rules to the new language.

\section{Understanding the semantics of data}
\label{data_section}
There is now a wealth of research in understanding the semantics of data, which is a crucial first step for semantic data science.  Broadly, the research in this area can be categorized into: (a) {\em table understanding}, which is to map cell values, columns, and indeed entire tables into a well defined set of concepts drawn from ontologies or knowledge graphs; (b) {\em inter-table data discovery}, which discovers for instance, other tables that can be joined or unioned based on a larger index of tables.  We describe the two separately below.

\subsection{Table Understanding}
Table understanding helps with explainability, trust, bias and feature engineering aspects of data science because it helps pick more suitable features overall to build models.  Here we cover a sample of the literature on table understanding to show that while challenging, tabular data understanding is possible.

\subsubsection{Knowledge graph based approaches}
Most approaches to table understanding assume the existence of one or more knowledge graphs or ontologies that define the universe of concepts to understand tabular cells or columns.  Columns that contain largely numeric values, dates, social security numbers, etc., are unlikely to appear in knowledge graphs, and are often called L columns.  For columns that contain strings which may potentially map to entities (called NE-columns), tabular understanding is broken into cell entity annotation (CEA), column type annotation (CTA), and column property annotation (CPA), with CEA often being used by many systems to help CTA and CPA.  An additional task considered by many systems is the detection of the subject column (S) which identifies the core entity column, with other columns in the table representing properties of the core entity.  Numerous systems now target these tasks, spurred by the presence of a challenge for these tasks \cite{semtab2019}, with the base knowledge graph being DBpedia or Wikidata.  Example systems that perform one or more of these NE tasks include MantisTable \cite{DBLP:journals/fgcs/CremaschiPRS20}, ColNet \cite{city22932}, TableMiner+ \cite{10.1007/978-3-319-18818-8_25}, and Sherlock \cite{10.1145/3292500.3330993}. Most of these systems are based on neural network based meta-learning on embeddings of column vectors. These approaches are powerful in some cases when plenty of good training examples are available. However, in many cases they do not scale well over the number of column concepts, and particularly on numerical data and in cases where training data is scarce or noisy. $C^2$\cite{khurana2020semantic} presents a maximum likelihood ensemble approach over multiple sources of data such as knowledge bases and noisy webtables and is more scalable in the number of concepts. Some systems, such as Memei \cite{Takeoka_Oyamada_Nakadai_Okadome_2019} jointly train on the CTA task along with column-column relationships to be able to annotate L columns.  

While the approaches and tasks performed vary across systems (e.g., supervised or semi-supervised), in almost all cases, the matching of columns to knowledge graphs is challenging enough for realistic tables. At best, it provides limited support for entirely automated semantic feature engineering.  This is in part due to at least two key limitations of the knowledge based approach: (a) Recall of knowledge bases is severely limited.  In one estimate, only 3\% of the tables contained in the 3.5 billion HTML pages of the Common Crawl Web Corpus can be matched to
DBpedia \cite{10.1145/2872427.2883017}. (b) The approach is especially challenging for L columns, unless the S column detection is perfect and the column heading maps to a knowledge graph property using CPA.  Knowledge agnostic techniques reviewed next circumvent some of the limitations of knowledge graph based approaches for poor recall.

\subsubsection{Knowledge graph agnostic approaches}
Knowledge graph agnostic approaches use large corpora of publicly available tables as knowledge, with schema matching techniques being used to map to tables and columns across the corpus.  The challenge here is how to scale this computation to millions of annotations, to perform what is often termed in the literature as holistic schema matching. Various approaches have been taken to this problem including a binary integration strategy that matches iteratively to create a single vocabulary \cite{10.1145/27633.27634}, or vectorizing column values and using techniques such as locality sensitive hashing (LSH) to minimize the number of pairwise comparisons (\cite{10.1007/978-3-642-35176-1_4}), amongst others.  

More recently, there has been an attempt to vectorize tables from large corpora.  Inspired by work in language models such as BERT \cite{bert}, which are effective in encapsulating correlations between words in a sentence, some recent works approach the problem of assigning types to columns by building unsupervised BERT based language models for columns based on their values \cite{trabelsi2020semantic}, adding other intermediate column labels to provide context to improve annotation.  In \cite{trabelsi2020semantic}, they use 1.6 million tables from the WikiTables domain to build the model.  This approach of using a corpus as a reference knowledge base is clearly attractive because it can greatly alleviate the recall problem of knowledge graphs. The availability of large table corpora such as those gathered from the web\footnote{\url{http://webdatacommons.org/webtables/}}, Viznet\footnote{\url{https://github.com/mitmedialab/viznet}}, Data.gov\footnote{\url{https://www.data.gov/}} make this approach promising and feasible.


\subsection{Inter-table discovery}
While features can often be constructed based on a single table, it is more common in data science that the scientist joins or unions data across tables to compute a model.  Table similarity detection targets this space of data wrangling.  Table similarity detection based on a number of metrics such as the number of matching rows, number of matching columns, the amount of information gained by possibly combining related tables, etc., are used by systems such as Juneau to increase the number of possible features for feature engineering \cite{10.1145/3318464.3389726}.  The problem of finding interesting joins in a data lake is addressed by JOSIE, which performs an efficient top-k overlapping set similarity search over millions of tables \cite{10.1145/3299869.3300065}.  Joinable tables are ranked by the size of overlap to a queried key column.  JOSIE works on equi-joins.  PEXESO \cite{Dong2021EfficientJT} performs joins on columns with string values using word embeddings, because joinable columns often have slight mismatches in the string representation.  PEXESO is also evaluated for data enrichment; the system shows 1.9\% higher micro-F1 score and 10\% lower mean squared error on machine learning tasks. KAFE~\cite{galhotra2019automated} and S3D~\cite{s3d} create effective joins and unions with knowledge graphs and tables large data lakes by identifying semantic concepts of the given data with those of the database table columns.

\section{Semantic Data Science through Mining}
Assuming we have the technologies to find useful sources of data to model a target variable, automated feature engineering still requires the mining of formulas that allow us to construct meaningful features for a given target, as illustrated in the BMI, CFR or loan examples.  We describe two areas for such mining - one from code, and another from text.

\subsection{Data science through mining code}
\label{code_section}
Recent work has targeted mining code repositories to help the data scientist.  Wranglesearch\footnote{\url{https://www.josecambronero.com/pdf/wranglesearch-draft.pdf}} is a system that mines Kaggle notebooks using dynamic analysis to extract functions and expressions that may help with data wrangling, and is perhaps the most relevant work for automated feature engineering.  Others target the larger data science development space.  \cite{10.1145/3447548.3467455} for instance mines Kaggle notebooks and scripts to find what they call decision points, where alternative models or parameters might be considered, and provide the data science users with these alternatives, so that they might build more robust models.  SOAR \cite{9402016} uses embedding techniques on the documentation of API calls combined with programming by example type approaches to suggest refactoring of data science code from one API to another (e.g. from tensorflow to PyTorch).

\subsection{Mining expressions from text}
\label{nlp_expression_section}
Vast amounts of domain knowledge useful for data science is buried in text as well.  There has been work that targets the extraction of topics for mathematical equations \cite{Yasunaga_Lafferty_2019} by building joint neural models for the textual context around the equation along with the equation itself.  In a data science context, it would be more useful for searching relevant expressions to surface to a data scientist, or to extract features if the terms of the equation can be mapped to columns successfully.  \cite{10.1145/3366423.3380218} describe the use of modified information retrieval techniques on an extracted database of several million equations; they, for instance, find equations associated with {\em Jacobi Polynomial}, or complete equations.  Once again, from a data science perspective, \cite{10.1145/3366423.3380218} connect equations to textual topics, but its utility in an entirely automated feature engineering mechanism is still an open question.

A related area of work is with respect to mining causal relationships from text; causality often determines which features may be semantically meaningful to model a target variable.  Work related to automated causal extraction from text is rich (e.g., \cite{LI2019512}, \cite{hassanzadeh2019answering}). Extraction of such information from notebooks and domain specific corpora could help with feature selection.


\section{Down-Stream Impact within Data Science}
\label{misc_section}
Data Scientists spend a majority of their time on data preparation, including discovery, visualization, engineering, and cleaning~\cite{anaconda}. The goal of such data wrangling extends beyond data transformation for the purpose of meeting the requirements for consumption by learning algorithms.  The common assumption is that the better the data is prepared, the more successful and trustworthy is the resulting model at the end of the Data Science workflow. 
In the previous sections, we discussed the impact of semantics on automated data engineering; now we highlight how semantics influences the down-stream processes, including model-building and evaluation, trust, and explainability.

\subsection{Model Building and Evaluation} 
If the features in the input data are appropriately annotated with their corresponding concepts and relationships to other features, a model may leverage this information in multiple ways. For instance, in a neural network, features may be explicitly separated by disconnecting the relevant nodes if they do not share any semantic relationship. Moreover, feature selection can use this information to prevent irrelevant and random features or spurious correlations that are often found (e.g., a high correlation between milk consumption and number of suicides\footnote{https://www.tylervigen.com/spurious-correlations}) in data, to be used to build a model.

Semantics can also be used to point out potential shortcomings in the building, evaluation, and monitoring of a machine
learning model. 
Humans can infer ``data leakage'' through semantic interpretation of features. For instance, pregnancy status, the presence of ovarian cancer, or skin tone as a feature are very strong predictors of variables such as  \texttt{gender} or
\texttt{race}. When data is appropriately annotated with latent features of gender or race, it can be accordingly
evaluated -- for instance, stratified by the latent features. In addition, when the machine can ``understand'' the meaning of the target variable, it can try to discover latent features in the data and if possible, reveal them to the user, and also
suggest an evaluation scheme. Of course, this may not prevent cases where statistical patterns predict
specific features with astonishing accuracy, but humans cannot (e.g., medical chest X-rays predicting
race~\cite{readingrace}). Nevertheless, AI-based semantics can uncover what humans quickly infer. AI could be used to extract purely statistical relations as researchers express them in text~\cite{readingrace}, and include them into a machine interpretable form such that an automated system for data science can use it. 

\subsection{Trust and Bias}
Protected attributes in data such as \texttt{gender}, \texttt{age}, and \texttt{race} are often used implicitly. 
For now, all trust-based or bias detection approaches need explicit
labeling of protected features by the user. They can be fairly effective but are purely statistical methods (e.g.,~\cite{trustsurvey}). Consider
the following example: in court systems, ML is used to predict if an undertrial will be a repeating offender.
Even though \texttt{race} is not a direct input feature, a learning algorithm can derive race implicitly based on where someone lives
-- which is inappropriate. As mentioned earlier, once data is semantically annotated -- either manually
or automatically -- AI can leverage this information to help data scientists with such challenges.
Assuming automatically semantically annotated data and a list of protected
attributes such as race, gender, or age, it is possible to discover the existence of
a potential relationship to a protected attribute by resorting to semantic techniques such as
determining relationships through knowledge graphs. In this sense, semantics also extend the reach of existing trust algorithms by automatically indicating when and how to employ them.

\subsection{Explainability and Transparency}
Explainability is an essential aspect of  models in data science~\cite{explainabilitysurvey}. Semantically driven automation can provide a human
understandable explanation on how new features were derived such as in Figure~\ref{fig:example}. This is in contrast to most features learned in neural networks where considerable effort goes in reverse-engineering semantics~\cite{meaningofneuros}. 
We note that semantic annotation does not mean that the machine 
actually understands the concepts, and it can even make semantically wrong choices. However, such errors would be easily
verifiable by a user instead of altogether recovering meaning from models such as boosted tree or neural network. 
Even though the feature derivation examples presented here are somewhat simple, they already go beyond what state-of-the art AutoML can achieve. We believe that it is a good starting point for understanding more about the data and the model. 
Insights such as the general understanding of the given data and its connections to other data sources could be related to it can now be automated to a reasonable extent. 
For instance, a relatively straightforward approach of submitting
a set of concepts from a given data to a search engine and some automated analysis of the returned documents (such as word clouds) can reveal the broader
context of a dataset. This can be further leveraged by the analytics or data discovery programs.

Equipping a data science automation program itself
with data semantics can make the entire data science workflow considerably more transparent and
explainable. This not only includes the ML pipeline related properties like why some data was
included or excluded, and how and why it got transformed; it also sheds light on the data science
workflow itself in terms of the chosen evaluation scheme and how the pipeline was constructed.

\subsection{Human-Computer Interaction} While effective human-computer interaction is important for AutoML by itself (e.g. H2O\footnote{H2O: \url{https://h2o.ai}}, DataRobot\footnote{DataRobot: \url{https://datarobot.com}}), it is particularly relevant for the case of semantically driven automation. Current state-of-the-art systems are not capable of determining all concepts and relationships on data accurately and reliably. However, a system that a user can control and guide with the help of limited automation, can become a very effective tool for Data Scientist and Data Engineers. We have witnessed this trend in other aspects of automation in machine learning as well~\cite{wang2021much}. 

\section{Conclusion}
\label{conclusion_section}
Data semantics is at the heart of data science. Conventionally, the modeling process has relied on data scientists' or domain experts' ability to detect semantics in given data, and connect them with external data or knowledge of the world. With a recent push for automation of the data science process itself, many aspects are being successfully automated. In this paper, we reviewed a broad range of approaches on how to discover and to leverage semantics to help with the same.
At present, we lack the capability for general purpose common-sense reasoning or deep domain understanding. However, we show that even limited machine-consumable semantics can enable AI driven automation of understanding-based data science. 


\bibliographystyle{named}
\bibliography{ijcai22}

\end{document}